\documentclass[svgnames]{amia}
\usepackage[svgnames]{xcolor}
\usepackage{graphicx}
\usepackage[labelfont=bf]{caption}
\usepackage[superscript,nomove]{cite}
\usepackage{color}

\usepackage{bm}
\usepackage{latexsym}
\usepackage{amsmath}
\usepackage{amssymb}
\usepackage{tikz}
\usepackage{array,multirow}
\usepackage{tabularx}

\begin{document}

\title{Benchmarking Modern Named Entity Recognition Techniques\\for Free-text Health Record Deidentification}

\author{Abdullah Ahmed, B.S.$^{1}$, Adeel Abbasi, M.D.$^{1}$, Carsten Eickhoff, Ph.D.$^{1}$}

\institutes{
    $^1$Brown University, Providence, RI, United States \\
}

\maketitle

\noindent{\bf Abstract} 

\textit{Electronic Health Records (EHRs) have become the primary form of medical data-keeping across the United States. Federal law restricts the sharing of any EHR data that contains protected health information (PHI). De-identification, the process of identifying and removing all PHI, is crucial for making EHR data publicly available for scientific research. This project explores several deep learning-based named entity recognition (NER) methods to determine which method(s) perform better on the de-identification task. We trained and tested our models on the i2b2 training dataset, and qualitatively assessed their performance using EHR data collected from a local hospital. We found that 1) BiLSTM-CRF represents the best-performing encoder/decoder combination, 2) character-embeddings and CRFs tend to improve precision at the price of recall, and 3) transformers alone under-perform as context encoders. Future work focused on structuring medical text may improve the extraction of semantic and syntactic information for the purposes of EHR deidentification.}

\section*{Introduction}
A majority of medical practices across the United States have adopted Electronic Health Records (EHRs). Between 2008 and 2016, EHR use by office-based physicians has nearly doubled from 42\% to 86\% \cite{ehr-data} -- an increase largely attributable to the Federal Health Information Technology (IT) Strategic Plan of 2011 \cite{hitech,it_plan}. One of the goals of this plan is to allow data within EHRs to be leveraged for scientific research. The use of EHR data continues to be restricted by the Health Insurance Portability and Accountability Act (HIPAA), whose Privacy Rule limits the distribution of patients' \emph{protected health information} (PHI). Unrestricted research use of EHR data is only permissible once it is \emph{de-identified} - all PHI has been removed. Per the HIPAA Privacy Rule, health information may be deemed de-identified through one of two methods: 1) ``Expert Determination,'' a formal conclusion by a qualified expert that the risk of re-identification is very small, and 2) ``Safe Harbor,'' the removal of 18 specified individual identifiers (names; geographic subdivisions; dates; telephone numbers; vehicle identiﬁers; fax numbers; device identiﬁers and serial numbers; emails; URLs; Social Security Numbers; medical record numbers; IP addresses; biometric identiﬁers; health plan beneﬁciary numbers; full-face images; account numbers; certiﬁcate or license numbers; any other identifier, code, or characteristic). 

Manual de-identification is tedious and time-consuming\cite{expert_difficult}. Researchers in the Natural Language Processing (NLP) community have developed systems to automate ``Safe Harbor'' de-identification processes by scanning medical free text for PHI identifiers. End-to-end de-identification involves three steps: 1) locating PHI in free text, 2) classifying the PHI correctly, and 3) replacing the original PHI with realistic surrogates. Step (3) is beyond the scope of this study; for simplicity, we will use the term ``de-identification'' to refer only to steps (1) and (2). De-identification can be framed as a named entity recognition (NER) problem. Formally, given a sequence of input tokens  $s=\{w_i\}_{i=1}^n$, an NER system outputs a list of tuples $<I_s, I_e, t>$, each of which is a named entity in $s$ \cite{ner_survey}. $I_s$ represents the start token, $I_e$ represents the end token, and $t$ is the entity type. $t$ is drawn from the 18 HIPAA PHI identifiers. 

Automatic de-identification methods fall into four broad categories: rule-based, machine-learning, hybrid, and deep learning. Rule-based systems rely on pattern-matching of textual elements \cite{rule_based}. They are simple to implement, interpret, and modify, but they require laborious construction, lack generalizability to unseen data, and cannot handle slight variations in language or word forms (\textit{e.g.}, misspellings, abbreviations). Machine learning systems model the de-identification task as a sequence labeling problem: given an input of tokens $w_1, w_2, \ldots, w_n$, the system outputs label predictions $y_1, y_2, \ldots, y_n$. Traditional machine-learning algorithms can recognize complex patterns in the data not evident to the human reader\cite{ml_based}. However, they require an input of handcrafted numerical ``features'' that are often time-consuming to engineer, and not guaranteed to be generalizable to other medical corpora. Hybrid methods combine elements of machine learning and rule-based systems \cite{hybrid}. Although they outperform their constituent parts, they still suffer from a lack of generalizability and a need for manual feature engineering. 

\begin{figure*}[t]
    \centering
    \includegraphics[scale=1.5]{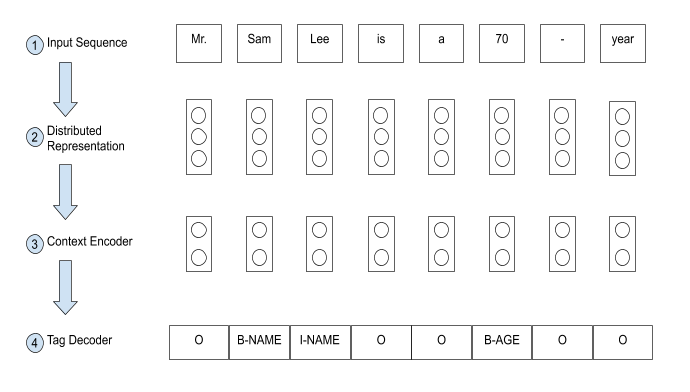}
    \caption{Pipeline taxonomy of deep-learning based NER systems.}
    \label{fig:dl_ner}
\end{figure*}

Deep learning -- a subset of machine learning based on artificial neural networks (ANNs) -- circumvents these problems. ANNs are capable of representation learning \textit{(i.e.}, automatically discovering useful features for a given task). In supervised learning, features are learned by training on a large set of labeled data of the form $(\bm{X},\bm{Y})$, where $\bm{X}$ and $\bm{Y}$ are the vector representations of the inputs and labels, respectively. Deep learning-based models can learn complex representations of token sequences through a series of non-linear transformations. Li \textit{et al.} \cite{ner_survey} outline the general structure of deep learning methods for NER, displayed in Figure~\ref{fig:dl_ner}. Once the tokenized sentence is passed into the model, it undergoes three stages of processing. The distributed representation stage converts every token to a numeric vector. The context encoder then processes these vectors to capture the contextual dependencies across the entire sentence, outputting a new sequence of vectors (not necessarily in the same dimensionality as the embeddings). Finally, the tag decoder uses the output of the context encoder to predict the label for each token. All deep learning-based NER systems can be characterized by the concrete design decisions made for each of these stages of processing. 

Recently, deep learning-based NER has been applied to de-identification \cite{dernocourt,liu_new,clinical_2}. Pre-trained word- and character-level embeddings have been employed in the first stage to form distributed representations of medical text. Recurrent Neural Networks (RNNs), specifically Long-Short-Term-Memory Networks (LSTMs), have demonstrated success in incorporating contextual information. Conditional Random Fields (CRFs) have gained popularity as a means of decoding the tags and predicting PHI labels in the final stage. In this study, we aimed to determine which NER design combinations perform better when tackling the de-identification task. Additionally, we aimed to extend the work of Yang \textit{et al.} \cite{clinical_2} and Yogarajan \textit{et al.} \cite{clinical_1} by evaluating the performance of our models on real EHR data collected from a local hospital.

\section*{Methods}

\begin{figure*}[t]
    \centering
    \includegraphics[scale=0.85]{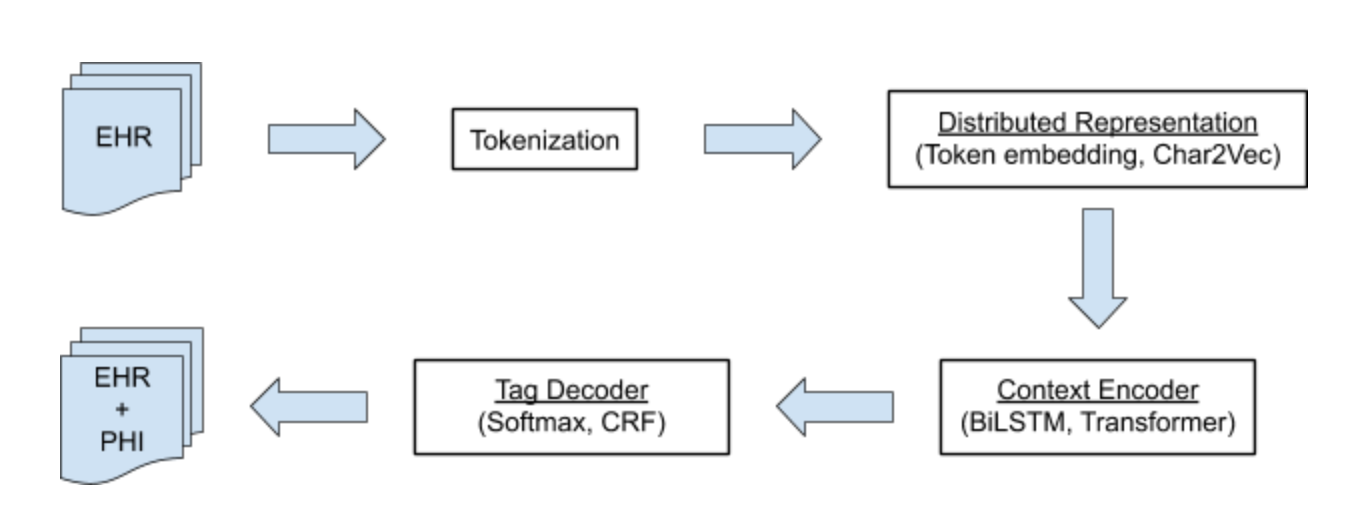}
    \caption{End-to-end pipeline of the NER design combinations we tested.}
    \label{fig:pipeline}
\end{figure*}

Figure \ref{fig:pipeline} summarizes the end-to-end structure of our system. The details are described in the following sections.

\noindent{\textbf{\textit{Data Collection}}}

Upon request, the Blavatnik Institute of Biomedical Informatics at Harvard University granted us access to the de-identification corpus created by the Center for Informatics for Integrating Biology and Bedside (i2b2) in 2014. The dataset contains 1,304 free-text medical records of patients with diabetes for which all PHI was manually annotated and replaced with surrogate PHI. This corpus was used for model training and quantitative evaluation.

Additionally, 25 health record notes collected between March and May of 2020 from Rhode Island Hospital (RIH) were included for qualitative manual performance inspection.  

\noindent{\textbf{\textit{Pre-processing}}}

Before feeding the medical text into our deep learning-based algorithms, we pre-processed it to take the form of sequences of sentences and tokens for each document. Formally, for every document $d \in D$, where $D$ is the set of all medical documents, we split $d$ into sentences $s_i$ and tokens $t_{i,j}$. 

Tokenization, the process of splitting sentences into tokens ($s_i \rightarrow t_{i,j})$, is a critical and highly customizable step for NLP systems. For many forms of free text, a tokenization scheme that splits based on whitespace and punctuation may suffice. However, the highly unstructured text in EHRs demands a more refined approach. We emulate the work of Liu \textit{et al.} \cite{liu_new}, which proposed a tokenization module that first splits on blank spaces, then recursively on other characters, words connected without a space, and numbers that appear adjacent to letters. For example, the EHR sentence ``Mr. SamLee is a 70yo man" would be tokenized as [`Mr.', `Sam', `Lee', `is', `a', `70', `yo', `man'].  This module preserves normally-occurring words and numbers whilst avoiding several pre-processing errors, such as ``SamLee" and ``70yo" in the example sentence above.

The set of all unique tokens in the training set is known as the ``vocabulary." The training and testing sets have equivalent vocabularies because the models are not permitted to incorporate any testing words into their training vocabulary. Any out-of-vocabulary (OOV) tokens -- words that appear in the testing set but not the training set -- were replaced with the \verb|UNK| token. This method allows models to generalize to tokens never encountered during training. 

For each token $t_{i,j}$, we also stored the characters it spans, $c_{i,j,s}$ and $c_{i,j,e}$, so that our results coincide with the i2b2 label format. Each sentence $s_i$ functions as a single training instance for our algorithms. Sentences were padded with \verb|PAD| tokens so that every sentence had the same length $m$. $\verb|PAD|$ tokens are masked during training loss calculation so that the model focused on predicting actual tokens correctly.

The last pre-processing step generated a sequence of labels for every sentence $s_i$, such that every token $t_{i,j}$ has a corresponding label $l_{i,j}$. We employed the popular BIO scheme to create the label sequence. Let $L_{i,s:e}$ be a PHI in sentence $i$ that starts at token number $s$ and ends at $e$. The BIO scheme prepends \verb|B-| (for beginning) to $l_{i,s}$ and \verb|I-| (for inside) to $l_{i,s+1:e}$. For instance, if ``Rhode Island Hospital'' appeared in sentence 2 and spanned tokens 14-16, the corresponding labels $l_{2,14:16}$ would be \verb|B-HOSPITAL,I-HOSPITAL,I-HOSPITAL|. Any tokens that do not qualify as PHI are assigned the label \verb|O| (for ``outside''). Figure~\ref{fig:dl_ner} demonstrates BIO tagging for a sample sequence.

\noindent{\textbf{\textit{Distributed Representation}}}

We leveraged information about the input across two levels (word and character) to form the embeddings for each token. At the word level, a token is viewed as a standalone unit. Every token in the training vocabulary was mapped to a unique vector in $\mathbb{R}^d$. The vectors were initialized to random values, and through training converged to useful representations. A random seed was set at the beginning of the program to ensure that the random vectors were initialized to the same values for every model we tested. At the character level, a token is viewed as a sequence of characters, allowing the model to incorporate sub-token patterns into the representation and thus capture additional semantic information from OOV tokens. Because it was trained on substantially more text than is available in the i2b2 dataset, we utilized a pre-trained character embedding layer, char2vec, to generate character-based embeddings. char2vec, trained using a Bidirectional Long-Short-Term-Memory (Bi-LSTM) to detect similar words based on character information \cite{char2vec}, outputs vectors in $\mathbb{R}^{50}$, which we concatenated with the token-level vectors in $\mathbb{R}^d$ to form a new distributed representation of each token. 

The use of pretrained word embeddings has led to dramatic successes in a wide range of NLP tasks. Pretrained word embeddings are embeddings learned through one task -- generally one that requires no labeled data -- and applied to solve a different task. While pretrained embeddings would likely have increased our models' performances, our study was focused on the foundational architectural components of the NER pipeline. We refer the reader to other work that focuses specifically on how pre-trained word embeddings improve performance on the de-identification task \cite{bert_deid}.

\noindent{\textbf{\textit{Context Encoder}}}

Recurrent Neural Networks (RNNs) have demonstrated success in capturing contextual information from variable-length sequential data. Let $x_1, x_2, \ldots, x_m$ be a sequence of vectors at steps $t = 1, \ldots, m$. In our task, $x_1, \ldots, x_m$ correspond to the distributed representations of each token in a sentence. Unlike a normal feed-forward network, RNNs maintain a hidden state $h_t$ that is fed as input into the model at time $t+1$ along with $x_{t+1}$. This way, the model is able to propagate prior information forward as the embeddings are sequentially processed.

RNNs lack the ability to capture long-term dependencies and suffer from the vanishing gradient problem. LSTMs attempt to alleviate these issues by incorporating a ``cell state" $c_t$. $c_t$ serves as a memory block that retains relevant information and discards irrelevant information collected up to time $t$. It does so through the use of forget and input gates; the forget gate controls what information from previous timesteps should be removed from the cell state, while the input gate controls what information from the current timestep should be added to the cell state. Furthermore, an output gate combines information from the current cell state $c_t$ and the previous hidden state $h_{t-1}$ to calculate a new hidden state $h_t$ (recall that RNNs only utilize $h_{t-1}$). Both $c_t$ and $h_t$ are transmitted to the next timestep for use in processing input $x_{t+1}$. 

Bi-LSTMs improve upon LSTMs by performing the same calculations in the reverse direction (with different parameters), thereby propagating contextual information in both directions. The final output of a Bi-LSTM is the concatenation of the hidden states from both the forward and backward passes. Bi-LSTMs are widely used in state-of-the-art NER systems, including those designed for de-identification \cite{dernocourt,liu_new, clinical_2}. Still, sustaining long-range dependencies is a challenge for sequential models such as Bi-LSTMs. In addition, because Bi-LSTMs perform sequential operations, they cannot be parallelized. These issues can be addressed by an alternative context encoder: a transformer. Transformer models have been adopted as context encoders for many NLP tasks, including NER \cite{devlin2018bert}, but are yet to be tested on the de-identification task. 

Transformers gained immense popularity in the NLP community following recognition of the power of self-attention in sequence-to-sequence (seq2seq) modeling (\textit{e.g.}, machine translation) \cite{attention}. Self-attention mechanisms simultaneously relate elements in a sequence to each other. Formally, attention is mapping of a query and a set of key-value pairs to an output, calculated as follows:
\begin{align*}
    Attention(Q,K,V) = softmax(\frac{QK^T}{\sqrt{d_k}})V
\end{align*}
where $Q, K, V$ are matrices of the query, key, and value vectors packed together, and $d_k$ is the dimension of the key vectors. $Q,K,V$ are calculated by multiplying the input sequence $x_1, x_2, \ldots, x_m$ by weight matrices $W_Q, W_K, W_V$ that are learned through training. 

Since transformers do not rely on sequential processing in their calculations, they have no inherent notion of token order. In order for the model to leverage positional information of the input tokens, positional encodings are added to the embedding of each element in the input sequence. The encoding function is designed such that the same token will have slightly different embeddings depending on where it appears in the sentence, thereby ``encoding" its position. We employed the same positional encoding function used by Vaswani\textit{ et al.} \cite{attention}.

We employed Multi-Head Attention by computing attention multiple times (with different $W_Q, W_K, W_V$), concatenating the results, and multiplying by another weight matrix $W_O$. The output of Multi-Head Attention is passed through a feed-forward network to retrieve the final output sequence. The transformer model proposed by Vaswani\textit{ et al.} \cite{attention} includes an encoder and a decoder, each consisting of several Multi-Head Attention ``blocks.'' Because the model was designed for tasks such as translation between languages, the decoder does not necessarily output sequences of length $m$ as necessitated by NER (one label for each token). Therefore, only the encoder portion of the model was used to encode context, retaining the benefits of self-attention and parallelization. 

\noindent{\textbf{\textit{Tag Decoder}}}

The tag decoder takes the output of the context encoder as input and produces a final sequence of tags. Sequence labeling can be cast as a multi-class classification problem; that is, for every token, output a probability distribution over all possible PHI (after BIO conversion). This can be achieved using a time-distributed dense layer with softmax activation. The dense layer is applied to each token, and the softmax activation creates a probability distribution over all the PHI for that token. For a vector $x$, softmax is calculated as follows: 
\begin{align*}
    p(y = j|x) = \frac{e^{(w_j^Tx + b_j)}}{\sum_{k\in K}e^{(w_k^Tx + b_k)}}
\end{align*}
where $w,b$ denote the weights and biases of the dense layer, $j$ is the index of one label, and $K$ is the set of all labels. To find the most probable label, we take the $argmax$ of the above equation. Because softmax assumes the tags to be independent, the probability of an entire sequence of tags $y_1, \ldots, y_m$ is given by
\begin{align*}
    p(y_1, \ldots, y_m|x) = p(y_1|x) \cdot \ldots \cdot p(y_m|x)
\end{align*}
A shortcoming of using the softmax approach is that every token and label is decoded independently, rendering it unable to capture patterns in the sequence of tags (\textit{e.g.}, \verb|I-HOSPITAL| is likely to follow \verb|B-HOSPITAL|). CRFs improve this by modeling dependencies between labels through graphical connections. In particular, linear-chain CRFs implement strictly sequential dependencies, as is the case in NER. Linear-chain CRFs define a global score for a sequence of tags as
\begin{align*}
    &C(y_1,\ldots,y_m|s_1,\ldots,s_m) = \\
    &b[y_1] + \sum_{t=1}^m s_t[y_t] + \sum_{t=1}^m T[y_t,y_{t+1}] + e[y_m]
\end{align*}
where $m$ is the length of the sequence, $T$ is a transition matrix between all tags, and $b,e$ are vectors that indicate the cost of beginning or ending on a given tag.  The scores $s_1, \ldots, s_m$ are obtained by passing the output of the context encoder through a linear dense layer of size $|K|$. $T$ contains parameters that encode how likely it is transition from one tag to the next, thereby capturing common sequences of tags that appear in the training data. Similar to softmax, CRFs model the posterior probability of a tag sequence using the following equation: 
\begin{align*}
\begin{split}
    p(y_1, \ldots, y_m = j_1, \ldots, j_m|s_1, \ldots, s_m) &=  \\ \frac{e^{C(j_1,\ldots,j_m|s_1,\ldots,s_m)}}{\sum_{k_1,\ldots,k_m \in K^m}e^{C(k_1 \ldots, k_m|s_1,\ldots,s_m)}}
\end{split}
\end{align*}
Linear-chain CRFs satisfy the optimal substructure property. Consequently, the calculations over possible sequences of tags can be completed efficiently via dynamic programming. The optimal sequence can be calculated using the Viterbi algorithm.

\newpage

\noindent{\textbf{\textit{Training}}}

Our networks were trained using cross-entropy loss, defined as 
\begin{align*}
    L = -\sum_i log(P({y_i}))
\end{align*}
where $y_i = y_{1i}, \ldots y_{mi}$ is the correct sequence of tags for sentence $i$. The probability $P$ is given by the outputs of the softmax and CRF decoders.

Adam has been shown to yield the highest performance and fastest convergence on sequence labeling tasks \cite{adam_ner}. Thus, we used the Adam optimizer with a learning rate of $0.001$ to update the network weights in batches of size 32 for 10 epochs.

\noindent{\textbf{\textit{Experiments and evaluation}}}

Table~\ref{models} lists the combinations of model components we tested. Model hyperparameters were selected according to the literature and constrained by GPU memory allocation. We trained each model independently on the official i2b2 training set and subsequently tested it on the official test set. All models were built using Tensorflow, a deep learning framework developed by Google.

\vspace{5pt}
\begin{table*}[h]
\centering
\caption{List of the tested models with their combination of representation, context encoder, and tag decoder.}
\begin{tabular}{|l|c c |c c|c c|}
\hline
\multirow{2}{*}{Model Name} & \multicolumn{2}{c|}{Distributed Repr.} & \multicolumn{2}{c|}{Context Encoder} &\multicolumn{2}{c|}{Tag Decoder} \\ \cline{2-7}
    & Token & Char2Vec & BiLSTM & Transformer & Softmax & CRF\\
\hline
BiLSTM & \checkmark && \checkmark && \checkmark & \\\hline
BiLSTM-CRF & \checkmark && \checkmark &&& \checkmark \\\hline
C2V-BiLSTM-CRF & \checkmark & \checkmark & \checkmark &&& \checkmark  \\\hline
Transformer & \checkmark &&& \checkmark & \checkmark &  \\\hline
Transformer-CRF & \checkmark &&& \checkmark && \checkmark  \\\hline
Transformer-BiLSTM & \checkmark && \checkmark & \checkmark & \checkmark & \\\hline
\end{tabular}
\label{models}
\end{table*}

To assess the performance of our models, we computed precision (PPV), recall (sensitivity), and $F_1$ of the PHI \textit{entities}. We evaluated entities rather than tokens because unidentified tokens represent an infringement of the HIPAA Privacy Rule. For the same reason, we used the i2b2 ``strict'' measure that only takes a prediction to be correct if the entire entity is matched exactly.

Let $\mathit{TP}$ stand for true positives, $\mathit{FP}$ stand for false positives, and $\mathit{FN}$ stand for false negatives. Precision calculates the proportion of correctly labeled PHI entities in the set of all PHI entities returned by the system (i.e. $\frac{TP}{TP+FP}$). Recall calculates the proportion of correctly labeled PHI entities in the set of all PHI entities in the test set (i.e. $\frac{TP}{TP+FN}$). $F_1$ is the harmonic mean of precision and recall (i.e. $2\cdot \frac{precision \cdot recall}{precision + recall}$). The overall performance of each system was evaluated using ``micro'' and ``macro'' versions of these metrics. Micro-average calculates metrics at the corpus level, whereas macro calculates them at the document level and averages the result over all documents. Furthermore, we calculated precision, recall, and $F_1$ for each HIPAA-PHI type; these metrics are reported on the token level to offer a more detailed insight into performance variation across different PHI types. All of these calculations were executed using the official i2b2 evaluation script.

To evaluate the generalizability of our model, we qualitatively inspected the results of our best system on EHR data collected from RIH between March and May of 2020. We were unable to perform quantitative analyses of the RIH data because it was not accompanied by any true PHI labels. 

\section*{Results}

Table~\ref{summary_statistics} shows descriptive statistics about the dataset after pre-processing. Table~\ref{global_results} displays the global results of our systems, evaluated at the macro-average level.  BiLSTM-CRF is the best-performing system according to all three metrics (0.8391, 0.818, 0.8284), followed closely by BiLSTM (0.8154, 0.7949, 0.805). 

\begin{table}[h]
\centering
\caption{
Summary statistics of the data after pre-processing
}
\begin{tabular}{|c | c c|}
\hline
 & Training & Testing \\
\hline
Sentences & 31,535 & 21,670 \\
Vocab Size & 23,905 & 23,905 \\ 
Tokens & 627,208 & 421,839 \\
PHIs & 15,953 & 10,834 \\
PHI Tokens & 44,298 & 30,006 \\
\hline
\end{tabular}
\label{summary_statistics}
\end{table}

\begin{table*}[h]
\vspace{5pt}
\centering
\caption{
Global performance (all PHI categories) on the test set. Metrics are reported on the macro-average level. 
}
\begin{tabular}{|c | c c c|}
\hline
Model & Precision & Recall & $F_1$ \\
\hline
BiLSTM & 0.8154 & 0.7949 & 0.805 \\
BiLSTM-CRF & \textbf{0.8391} & \textbf{0.818} & \textbf{0.8284} \\ 
C2V-BiLSTM-CRF & 0.7925 & 0.3183 & 0.4542 \\
Transformer & 0.5027 & 0.6345 & 0.561 \\
Transformer-CRF & 0.6068 & 0.5843 & 0.5953 \\
Transformer-BiLSTM & 0.7259 & 0.6865 & 0.7056 \\
\hline
\end{tabular}
\label{global_results}
\end{table*}

Table~\ref{phi_results} lists the performances of our models on HIPAA-PHI categories, evaluated at the micro-level.  BiLSTM-CRF has the highest $F_1$ score in all categories. The addition of a transformer to the context encoder in Transformer-BiLSTM improved precision in both the AGE and CONTACT categories. Char2vec significantly improved the precision of LOCATION and slightly improved the precision of DATE, yet suffered tremendously in recall as a consequence. BiLSTM had higher recall than BiLSTM-CRF in three categories. DATE yielded the highest scores in all model combinations, owing to the constant, structured format in the i2b2 dataset (MM/DD/YYYY). 

\begin{table*}[h]
\centering
\vspace{0.3cm}
\caption{
Performance per category. Metrics are reported on the micro-average level.
}
\begin{tabular}{|l|c c c|c c c|c c c|}
\hline
\multirow{2}{*}{\parbox{2cm}{\centering PHI Category}} & \multicolumn{3}{c|}{BiLSTM} & \multicolumn{3}{c|}{BiLSTM-CRF} &\multicolumn{3}{c|}{C2V-BiLSTM-CRF} \\ \cline{2-10}
    & P & R & $F_1$ & P & R & $F_1$ & P & R & $F_1$\\
\hline
NAME & 0.9012 & 0.7238 & 0.8028 & \textbf{0.9268} & \textbf{0.7266 }& \textbf{0.8146 } & 0.8992 & 0.2482 & 0.389 \\
PROFESSION & 0.7331 & 0.5389 & 0.6212 & \textbf{0.8148} & 0.5483 & \textbf{0.6555} & 0 & 0 & 0
 \\ 
LOCATION & 0.7792 &	\textbf{0.6181} & 0.6894 & 0.7975 & 0.6174 & \textbf{0.696} & \textbf{0.8814} & 0.2681 & 0.4112
 \\
AGE & 0.8863 & \textbf{0.9241} & 0.9048 & 0.9407 & 0.8868 & \textbf{0.913} & 0.8843 & 0.2543 & 0.395
 \\
DATE & 0.9798 & 0.9675 & 0.9736 & 0.9703 & \textbf{0.9839 }& \textbf{0.9771} & \textbf{0.9914} & 0.228 & 0.3707
 \\
CONTACT & 0.6416 & \textbf{0.619} & 0.6301 & 0.8226 & 0.5113 & \textbf{0.6306} & 0.7619 & 0.401 & 0.5255
 \\
ID & \textbf{0.8943} & 0.7099 & 0.7915 & 0.8398 & \textbf{0.7847} & \textbf{0.8113} & 0.8631 & 0.4142 & 0.5598
\\
\hline
\multirow{2}{*}{} & \multicolumn{3}{c|}{Transformer} & \multicolumn{3}{c|}{Transformer-CRF} &\multicolumn{3}{c|}{Transformer-BiLSTM} \\ \cline{2-10}
    & P & R & $F_1$ & P & R & $F_1$ & P & R & $F_1$\\
\hline
NAME & 0.6922 &	0.7117 & 0.7018 & 0.7401 & 0.6074 & 0.6672 & 0.8171 & 0.5914 & 0.6862
 \\
PROFESSION & 0.4403	& \textbf{0.5514} & 0.4896 & 0.6959 & 0.4704 & 0.5613 & 0.6307 & 0.4735 & 0.5409
 \\ 
LOCATION & 0.6772 & 0.5137 & 0.5842 & 0.5987 & 0.4925 & 0.5404 & 0.6189 & 0.5571 & 0.5863
 \\
AGE & 0.7779 & 0.8628 & 0.8182 & 0.8176 & 0.8176 & 0.8176 & \textbf{0.9508} & 0.7976 & 0.8675
 \\
DATE & 0.8083 & 0.8564 & 0.8317 & 0.8563 & 0.824 & 0.8398 & 0.9603 & 0.9805 & 0.9703
 \\
CONTACT & 0.3953 & 0.2932 & 0.3367 & 0.4854 & 0.1253 & 0.1992 & \textbf{0.8525} & 0.2607 & 0.3992
 \\
ID & 0.6443 & 0.5967 & 0.6196 & 0.6751 & 0.4626 & 0.549 & 0.8887 & 0.4443 & 0.5925
\\
\hline
\end{tabular}
\label{phi_results}
\end{table*}

\begin{table*}[t]
\vspace{5pt}
\centering
\caption{
Output of BiLSTM-CRF \textit{vs.}\ C2V-BiLSTM-CRF on a sentence that contains several OOV terms. Character embeddings are able to identify ``HESS'' and ``CLARENCE'' as PATIENT tokens, whereas BiLSTM-CRF is not.
}
\begin{tabular}{|l | c c c c|}
\hline
Original Tokens & HESS & , & CLARENCE & 64365595 \\
Test Tokens & UNK & , & UNK & UNK \\ \hline
BiLSTM-CRF & B-HOSPITAL & O & O & O \\
C2V-BiLSTM-CRF & B-PATIENT & O & I-PATIENT & O \\ 
True Labels & B-PATIENT & I-PATIENT & I-PATIENT & B-MEDICALRECORD \\
\hline
\end{tabular}
\label{example}
\end{table*}

Table~\ref{example} highlights one example in which Char2vec improved the ability to predict the label for OOV tokens. Table~\ref{covidehr_1} shows the results of our BiLSTM-CRF model on ten samples of EHR data collected from RIH. The samples show that although our model generalizes to some pieces of PHI, it struggles with others that are unlike the ones present in the i2b2 dataset (\textit{e.g.}, signature formats).

\section*{Discussion}

In this study, we tested several different combinations of NER components -- distributed representations, context encoders, and tag decoders -- for EHR de-identification. We found that BiLSTM-CRF, introduced by Huang \textit{et al.} \cite{huang} for general NER outside of the clinical domain, is the best overall encoder/decoder combination for de-identification. Our results are in agreement with Dernoncourt \textit{et al.} \cite{dernocourt}, Liu \textit{et al.} \cite{liu_new}, and Yang \textit{et al.} \cite{clinical_2}.

Despite BiLSTMS-CRF's overall superior performance, Table~\ref{phi_results} shows that other configurations can locally outperform BiLSTM-CRF for some of the HIPAA-PHI categories. We attribute these findings to the distribution patterns of tokens in each category. LOCATION, for example, includes ZIP codes, sequences of five numbers. Char2vec is able to recognize that ZIP codes are consistently tokens with a length of five and composed only of numbers. Therefore, C2V-BiLSTM-CRF is the model most equipped to classify ZIP codes, contributing in part to its nearly 10\% increase in LOCATION precision. 

Furthermore, we found that character embeddings improved precision in several categories, yet decreased recall. This implies that morphological information captured by character embeddings increased the model's accuracy in identifying PHI type, yet decreased its sensitivity in detecting PHI. Disambiguation of PHI type is especially difficult for OOV tokens without the use of character embeddings, as evidenced in Table~\ref{example}. The decline in recall is likely because the char2vec embeddings were not fine-tuned during the training process. Thus, the character embeddings remained static, unable to adapt to the distribution of medical text. Alternative character embeddings, such as those that utilize Convolution Neural Networks (CNNs) \cite{cnn_1}, could also improve performance. 

In our study, transformers were less effective than Bi-LSTMs at encoding context. This may be accounted for by the uncontrolled sentence lengths in EHRs. Due to the transcriptional style of medical text, there are ``sentences'' that contain over a thousand tokens ($m$ = 1567). As a result, the transformer model may try to capture long-term dependencies via self-attention in the absence of meaningful relationships. Moreover, in an analysis of encoder representations in transformers, Raganato \textit{et al.} \cite{raganato2018analysis} show that syntactic information is captured in the first 3 layers of the encoder, while semantic information is captured later. The transformer we used, which only had two layers of multi-headed attention, may have only partially captured the syntactic information of a distribution of medical text that conformed to limited syntactic rules. Performance improved when a Bi-LSTM was stacked on top of the transformer, potentially having compensated for the lack of captured semantic information. Future research may explore adding more transformer layers to the context encoder to extract more semantic information. 

While CRFs as tag decoders generally improve the $F_1$ score, our results show that they can decrease recall for both Bi-LSTM and transformer context encoders. We hypothesize that certain sequences of tags seen in the training set became favored by the model, leading to unseen sequences in the testing set receiving low likelihoods. 

A hybrid method that leverages the strengths of each model -- based on its performance in individual PHI categories -- may function best in practice. For instance, BiLSTM-CRF could be used to output an initial set of candidate PHI's because it has the highest $F_1$ score in all categories. The candidates could then be filtered using models with high specificity scores, such as Transformer-BiLSTM for AGE and CONTACT predictions and C2V-BiLSTM-CRF for LOCATION and DATE predictions. 

Qualitative assessment of our top model with the EHR data collected from RIH indicates that it somewhat generalizes beyond the i2b2 dataset (Table \ref{covidehr_1}). It was still able to classify crucial PHI such as Medical Record Numbers (MRNs), account numbers, and dates. However, it failed with sentences and phrases whose formatting significantly differs from i2b2 (\textit{e.g.}, signature formats, incomplete phone numbers), as well as with tokens it never encountered (e.g. ``COVID-19"). One particularly revealing example is the classification of ``Rhode Island Hospital" vs. ``RIH ER." Our model could correctly classify the former because it extrapolated from similar hospital names it encountered during training. On the other hand, it was unable to extract any semantic information from the abbreviated form and thus misclassified it. 

To alleviate the problem of model portability, Yang \textit{et al.} \cite{clinical_2} show that fine-tuning their model on labeled data from local hospital EHRs improves their performance. We were unable to do the same because the EHR data we received contained no PHI labels. Future research might explore the utilization of local EHR data to fine-tune a language model that is independent of the de-identification pipeline, drawing inspiration from models like ClinicalBERT that are fine-tuned on clinical text \cite{clinical_bert}. That said, recent research has shown that it is possible to extract personally identifiable information from large language models through adversarial attacks \cite{carlini2020extracting}. More work must be done to protect against these attacks before safely incorporating PHI into training data. 

The underlying problem remains that medical text is highly unstructured and non-standardized, resulting in sentences that lack syntactic and semantic cohesiveness. Without structured information, it becomes near impossible to automatically achieve results that fully satisfy the HIPAA Privacy Rule and are portable to multiple hospital systems. At the lowest level, the text must be tokenized in a way that permits inference. Dedicated medical tokenizers like Medex exploit domain knowledge to extract information about medications from medical narratives \cite{medex}. However, this does not resolve the long and disorganized nature of medical text. Recent efforts to enforce structure upon notes using NLP may help in downstream tasks like de-identification that rely on extracting very specific information \cite{narr_struct}. Uniformity in note structure will not only improve the model's performance but will also increase its ability to generalize beyond the data used in training. 

\section*{Conclusions}

\begin{table*}[t]
\centering
\caption{
Sample output of BiLSTM-CRF for phrases in the RIH EHR dataset. Original tokens have been manually replaced. \textcolor{Green}{Green} entities are PHI correctly identified ($TP$), \textcolor{red}{red} entities are PHI that went unidentified ($FN$), and \textcolor{orange}{orange} entities were incorrectly identified as PHI ($FP$). 
}
\begin{tabularx}{11cm}{|X|}
\hline
Mr. \textcolor{Green}{Smith} is a \textcolor{Green}{200}-year-old gentleman \\ \hline
Admitted to \textcolor{Green}{Rhode Island Hospital} for \textcolor{orange}{COVID-19} \\ \hline
travel to \textcolor{Green}{YZ} from \textcolor{Green}{8/20} - \textcolor{Green}{8/26} who presented to \textcolor{red}{RIH ER} on \textcolor{Green}{8/28/60} \\ \hline
Signature: \textcolor{red}{Sam Lee}, MD \textcolor{orange}{Electronic Signature} \\
\hline
I communicated with this patient's father \textcolor{red}{John Smith} at \textcolor{Green}{123-456-7890}\\\hline
\textcolor{Green}{Sunday} will be the last day of therapy \\ \hline
Vent Mode: \textcolor{orange}{PC FiO}2 (\%) [50\% - 100\%] \\ \hline
until \textcolor{Green}{Sunday}, as discussed with dr. \textcolor{Green}{Lee} \\ \hline
Social work will continue to follow. LICSW \textcolor{red}{456}-\textcolor{Green}{7890} \\ \hline
MR \#: \textcolor{Green}{00000000000} Account \#: \textcolor{Green}{111111111} \\ \hline
\end{tabularx}
\label{covidehr_1}
\end{table*}

This study gives a comprehensive review of wide-ranging information extraction techniques on the de-identification of EHRs. Through empirical testing of different NER design combinations, we found that BiLSTM-CRF is the best-performing encoder/decoder combination for the de-identification task. Character-embeddings and CRFs tend to improve precision at the cost of recall. Meanwhile, transformers alone underperformed as context encoders. Qualitative assessment of BiLSTM-CRF on local EHR data showed some success, yet the issue of model portability remains. Future work lies in automatically structuring medical text such that semantic and syntactic information can more easily be extracted and models become more generalizable.

\makeatletter
\renewcommand{\@biblabel}[1]{\hfill #1.}
\makeatother

\bibliographystyle{vancouver}
\bibliography{ref.bib}

\end{document}